# Enaction-Based Artificial Intelligence: Toward Co-evolution with Humans in the Loop.

Pierre De Loor[1], Kristen Manac'h and Jacques Tisseau

**Draft - Paper**

**Abstract** This article deals with the links between the enaction paradigm and artificial intelligence. Enaction is considered a metaphor for artificial intelligence, as a number of the notions which it deals with are deemed incompatible with the phenomenal field of the virtual. After explaining this stance, we shall review previous works regarding this issue in terms of artifical life and robotics. We shall focus on the lack of recognition of co-evolution at the heart of these approaches. We propose to explicitly integrate the evolution of the environment into our approach in order to refine the ontogenesis of the artificial system, and to compare it with the enaction paradigm. The growing complexity of the ontogenetic mechanisms to be activated can therefore be compensated by an interactive guidance system emanating from the environment. This proposition does not however resolve that of the relevance of the meaning created by the machine (sense-making). Such reflections lead us to integrate human interaction into this environment in order to construct relevant meaning in terms of participative artificial intelligence. This raises a number of questions with regards to setting up an enactive interaction. The article concludes by exploring a number of issues, thereby enabling us to associate current approaches with the principles of morphogenesis, guidance, the phenomenology of interactions and the use of minimal enactive interfaces in setting up experiments which will deal with the problem of artificial intelligence in a variety of enaction-based ways.

# Introduction

Over the past few years, the cognitive sciences have been undergoing considerable evolution having taken into account the natural and committed nature of organisms when describing their cognitive capacities (Sharkey & Ziemke, 1998; Lakoff & Johnson, 1999). Enaction is one of the theoretical propositions involved in this evolution (Varela, Thompson, & Rosch, 1993; Noë, 2004; Stewart, Gapenne, & E. Di Paolo, 2008). Even though debate about the

---

[1] UEB - ENIB - LISyC, Centre Européen de Réalité Virtuelle, Brest, 29200, France

Email : deloor@enib.fr, manach@enib.fr, tisseau@enib.fr

www.enib.fr/~deloor, www.cerv.fr

relevance of the different areas of the cognitive sciences seems to be quieting (Gershenson, 2004), enaction offers an alternative to cognitivism (Pylyshyn,1984) and connectionist approaches (Rosenblatt, 1958) by following and furthering the sensorimotor theories initiated by (Gibson, 1966). It is based on research in the fields of biology (Maturana, Uribe, & Frenk, 1968; Maturana & Varela, 1980) and neuroscience (Freeman & Sharkda, 1990; Freeman, 2001). It supports constructivism (Piaget, 1970; Foerster, 1984; Shanon, 1993; Glasersfeld,1995; Rosch, 1999) and anthropological argumentation (Hutchins, 2005, 2006). Finally, its philosophical extension is also reiterated in phenomenology (Husserl, 1960; Merleau-Ponty, 1945; Varela et al., 1993; Lenay, 1996; Bickhard, 2003) and is at the centre of the research program into neurophenomenology (Thompson & Varela, 2001; Lutz, Lachaux, Martinerie, & Varela, 2001). Enaction supports the construction of cognition on the basis of interactions between organisms and their physical and social environments (De Jaegher & Di Paolo, 2007). It is thus rooted in radical constructivism. The issue which we will be analyzing here is that of the links woven between enaction and artificial intelligence, first dealt with a few years ago.

Even if an auto-constructing artificial system is not in itself new to artificial intelligence (Turing, 1950; Von Neumann, 1966; Drescher, 1991; Hall, 2007), the Computational Theory of Mind faces a number of difficulties linked to the representational nature which it proposes (Dreyfus, 1979; Fodor, 2000)[2]: the frame problem (McCarthy, 1969; Korb, 2004), the symbol grounding problem (Harnad, 1990, 1993), modeling of common sense (McCarthy, 1969), the importance of context (Minsky, 1982; McCarthy & Buva, 1998), creativity or indeed social cognition, or cognition in an open environment. In order to overcome these difficulties, new AI rejects the idea of representations and is at the source of embodied-embedded AI (Brooks, 1991; Pfeifer & Gomez, 2005).This approach integrates the role of the body and the sensorimotor loop in recognizing a robot's cognitive capacities. Nevertheless, it encounters difficulties regarding questions of agentivity, teleology and construction of meaning. (Ziemke, 2001; Di Paolo, 2005; Di Paolo, Rohde, & De Jaegher, 2007) differentiate between automatic systems, which rely on fixed exterior values, and systems which create their own identity. The biological origins of these notions, predicted by I. Kant (Kant, 1790), J. von Uexküll (Uexküll, 1957) or H. Jonas (Jonas, 1968) seem to be one possible key element in resolving these issues. As such, one would need to meticulously copy natural mechanisms artificially (Dreyfus, 2007). A task of such complexity seems unfathomable, however (Di Paolo & Iizuka, 2008) insist that it is not the details of these mechanisms that count, but rather the underlying principles which much be identified. It is these principles which aim to clarify enaction via a radical point of view according to which, due to the viability constraints of organisms and on their capacity to react, their interactions "crystalize" the sensorimotor invariants which are thus the source of enacted "embodied representations" from agentivity and from sense-making (Di Paolo, 2005). The paradigm demands an absence of representations of a pre-given world and also of the biological origins of autonomy: the autopoiesis principle. This principle is extended further by the integration of the sensorimotor loop, the co-evolution of the organism and its environment, and finally the enaction of its own-world. The notion of own-world (or phenomenal world) (Uexküll, 1957), expresses the way in which a subject's representation of the world is unique to that person and cannot be detached from his personal experience and sensorimotor capacities.

---

[2] These difficulties also relate to the connectionist approaches which, in this context, constitute a cognitive background, maintaining cognition at the status of a entrance/exit processing system.

In terms of the virtual, enaction and its revolutionary vision enable us to lay down new foundations. These new foundations led (Froese & Ziemke, 2009) to lay down the guidelines for "Enactive Artificial Intelligence" which clears the existing ambiguities surrounding the notion of embodied cognition highlighted by (Clark, 1999; Nunez, 1999; Ziemke, 2004). We will remain prudent about the terms we use, considering enaction as a metaphor for artificial intelligence. We shall therefore refer instead to "Enaction-Based Artifical Intelligence (EBAI)". Indeed, the direct transfer of a paradigm from the cognitive sciences might lead to shortcuts, misunderstanding and confusion regarding the initial notions of the paradigm. For example, enaction borrows the specificity of first-hand experience from phenomenology, and it is necessary to use phenomenology in order to understand the mind. However, in the case of machines, the notions of first-hand experience, consciousness and own-world are without a doubt inaccessible, if not absurd. This article does not aim to enter into the debate surrounding the functionalism of the intentionality of autonomy or of consciousness (Searle, 1997; Chalmer, 1995; Pylyshyn, 2003; Kosslyn, Thomson, & Ganis, 2006; Thompson, 2007). We shall simply embark on analyses of (Rohde & Stewart, 2008) who propose to replace the traditional distinction between ascriptionnal and genuine autonomy by presenting the hypothesis that "*an attributional judgement based on knowledge of an underlying behavior-inducing mechanism will be more stable than a naïve judgment based only on observation of behavior*". This concept enables us to use the ideas and advances of cognitive science in order to contribute to the artificial sciences (Simon, 1969) and vice versa. In particular, the problem of sense-making, crucial in artificial intelligence, can be established in an enactive inspiration.

This article will be structured in the following manner: section 2 outlines the notions relating to enaction and the characteristics expected of an artificial system claiming to adhere to the model. In section 3 we shall summarize the main elements of the approaches in artificial life and robotics which fall into the category of enaction. For each of these approaches, we will demonstrate how little importance is given to the evolution of the environment and the difficulties involved in obtaining ontogenetic mechanisms. The notion of a sense-making for a machine can also be a problem for a human user if it is designed to be autonomous in a purely virtual world. Having studied these issues, we make a number of suggestions in section 4: a more explicit recognition of the irreversible evolution of the environment and of coupling; guiding the artificial entity in order to tackle more complex ontogenesis as is the case in the co-evolving nature and integration of the "man-in-the-loop" with the co-creation of meaning, compatible with the social construction of meaning and the initial precepts of AI, illustrated by the Turing test. The section then goes on to present the areas which we shall explore in future research in order to meet these goals, before going on to the conclusion (section 5).

# From enaction to artificial intelligence

Enaction proposes to address cognition as the history of structural coupling between an organism and its environment. Here follows a brief summary of the concepts closely linked to it. For a more detailed account, we recommend the review articles by (McGee, 2005, 2006). Enaction originates from the notion of autopoietic systems put forward by Maturana and Varela as model of the living centred on the capacity of organisms to preserve their viability (Varela, Maturana, & Uribe, 1974). For these authors, this preservation defines the organism's autonomy and constitutes the biological origin of its cognitive capacities. An autopoietic system is a structure which produces itself as a result of its environment. The environment may disrupt the system, whose functioning will evolve as a consequence of that effect. If the functioning of the organism evolves in such a way as to preserve it despite disruption from exterior factors, the organism can be considered viable. This new way of functioning will, in return, influence the environment and the organism-environment system will co-evolve. The fact that the environment is but a disruption implies that it does not seem to be represented within the organism as a pre-given world. Furthermore, constraints on viability and the necessity to remain alive endow the organism with an identity by means of its metabolism and its capacity to act. This identity emerges relative to viability constraints, and the environment gradually takes on meaning.

Breaking away from biology, we talk about operationally closed systems. Operationally closed systems form a system of recursively interdependent processes in order to regenerate themselves, and can be identified as a recognizable unit in the domain of processes. Nothing prevents the notion of operationally closed systems being applied to the phenomenal domain of the artificial. The scientific approach would then be to generalize this mechanism to multicellular organisms (Varela, 1979), and thus to human beings, the mind, and social cognition (De Jaegher & Di Paolo, 2007). At each level, there is a difference linked to the aspects associated with the notions of viability and unity (Stewart, 1996; Di Paolo, 2005). Without entering into further detail and the arguments behind the theoretical approach, we shall retain three important characteristics involved in the development of artificial systems based on this paradigm:

1- The absence of a priori representations: In the domain of AI, this characteristic shares similarities with Rodney Brooks considerations (Brooks, 1991) but which, to be more precise, translates to an absence of representations of a pre-given world. The organism does not possess an explicit and definitive representation which it could manipulate in the manner of an imperative program, for example to plan or define an intention as a rule-based calculus. It is these interactions which enable it simply to "survive" by preserving sensorimotor invariants.

2- Plasticity: The organism is viable as it is capable of "absorbing" the disruptions caused by its environment and to adapt to them. This plasticity can be observed not only in the body for physical interactions but also at nerve level for higher-level interactions (cerebral plasticity).

3- Co-evolution: requires the distinction between physically grounded cognition and cognition that is rooted in their own world (Sharkey & Ziemke, 1998). A modification of the world by the organism in return imposes a modification of that organism. This co-evolution can just as well be considered a phylogenetic scale as an ontogenetic scale and gives rise to structural coupling characterized by its irreversibility. The example is often giving of tracing a path by trampling the ground with our feet.

In this way, we can see that the artificial system is taking the form of a complex system i.e. it is heterogeneous, with an open and multi-scaled dynamic (Laughlin, 2005). The emergent properties of these systems are testimony to the openness and the multiplicity of the possibilities of evolution. The notion of "natural derivation", highly important in enaction (Varela et al., 1993) is thus converted to "artificial derivation". It underlies complex systems and can initiate creativity and commitment in "bringing forth a new world". Creativity is here defined as the possibility to determine the functions of an undefined element of the environment.

These systems are able to apprehend and to enact properties relating to the world with which they interact. These properties, which are often dynamic, are difficult to represent using symbols and also resist abstraction. These characteristics are fundamental to enaction, which considers that know-how precedes knowledge and highlights the uniqueness of each experience.

Co-evolution involves a recursive transformation of the system and of its environment. The environment is thus an actor in the same way as the entity that occupies it. However, generally, the theories of embodied AI neglect the evolution of the environment, preferring to focus on perfecting the autonomous system. This priority is illustrated by the first "Enactive AI design principles" drawn up by (Froese & Ziemke, 2009):

- **principle EAI-1a:** an artificial agent must be capable of generating its own systemic identity at some level of description.
- **principle EAI-1b:** an artificial agent must be capable of changing its own systemic identity at some level of description.

Systemic identity works from the notion of auto-maintenance of a system as it is understood in the theory of autopoiesis. Principle 1b is a compromise made due to the complexity of implementing principle 1a. The second set of principles introduces the concept of interaction between the organism and the environment by means of the sensorimotor loop:

- **principle EAI-2a:** an artificial agent must be capable of generating its own sensorimotor identity at some level of description.
- **principle EAI-2b:** an artificial agent must be capable of changing its own sensorimotor identity at some level of description.

The active behavior of the agent is here dealt with explicitly. It enables us to address the construction of meaning in terms of a preservation of sensorimotor loops, but ignores the co-evolution of the environment and the agent. In conclusion, the role of the environment and its relative capacity to endanger the viability of the agent, is introduced by the third principle:

- **principle EAI-3:** an artificial agent must have the capacity to actively regulate its structural coupling in relation to a viability constraint.

However, to us, the irreversible nature of the conjoined evolution of the entity and its environment does not seem to have been made clear. For now the challenge is to introduce regulatory mechanisms in order to maintain the existence of the entity, knowing that the impositions exerted on it will evolve. The system must be able to regulate its regulation, to be able to access a meta-regulation (Morin, 1980). The implementation of such a system stems from (Froese & Ziemke, 2009) and particularly the hard problem of enactive artificial intelligence. This consists of concretizing the set of rules governing the system so as to define the modifications enabling it to be preserved. To do so would imply an understanding between the domain of explicit design and that of evolutionary approaches. This is the only method currently available when attempting to set up auto-adaptive artificial systems which rely on a dynamic rather than a representational approach. Before putting forward our suggestions for overcoming this problem, we shall identify the ways in which current approaches adhering to the artificial enaction paradigm fail to consider the role of the environment and of co-evolution in sufficient detail.

# Co-evolution and environnement in (enactive) artificial intelligence ?

Research corresponding to an enactive approach to artificial intelligence logically developed in the domain of artificial life alongside the study of the principles of autopoiesis (McMullin, 2004; Beer, 2004; Bourgine & Stewart, 2004; Beurier, Michel, & Ferber, 2006; Ruiz-Mirazo & Mavelli, 2008). These studies concern principles EAI-1a and EAI-1b. Other research in robotics has followed a similar trend with the development of artificial dynamic cognition which can be associated with the study of principles EAI-2b and EAI-3 (Beer, 2000; Di Paolo, 2000; Nolfi & Floreano, 2000; Harvey, Di Paolo, Wood, Quinn, & Tuci, 2005; Wood & Di Paolo, 2007; Iizuka & Di Paolo, 2007). We shall summarize these findings focusing particularly on the assimilation of environment and co-evolution.

**Simulating autopoiesis: The biological origins of autonomy**

*Principles*

The theory of enaction is rooted in the biological mechanism of autopoiesis. The autonomy of an autopoietic system constitutes its minimal cognition. We must remember that an autopoietic system is a composite unit, much like an element-producing network in which the elements 1) via their interactions, recursively regenerate the network of production which produced them and 2) construct a network in which they exist by building up a frontier with their external surroundings via their preferential interactions within the network (Dempster, 2000). Autopoietic systems possess the properties of emergent systems as they are able to create natural phenomena independent of those from which they were generated (Laughlin, 2005). Figure 1 summarizes the principles of minimum autopoietic systems models.

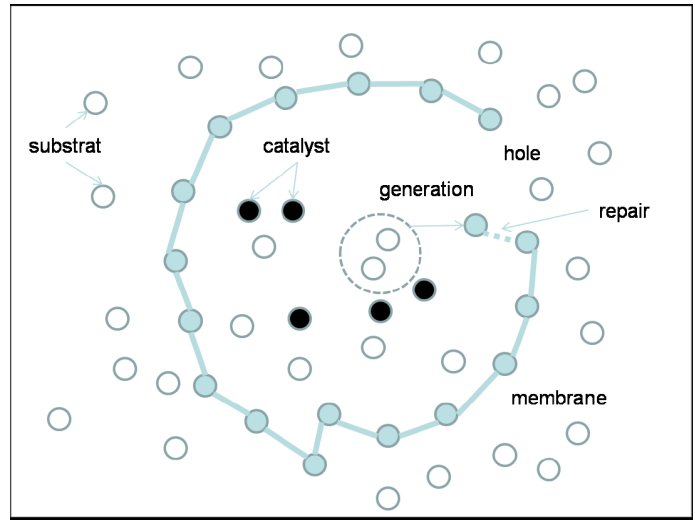

**Fig. 1** Illustration of the autopoiesis principle: A cellular membrane encloses a catalyst which cannot cross that membrane. A substrate can cross the membrane. In the presence of the catalyst, the substrate evolves into elements which will repair the membrane, should holes appear in it. Thus the cell is able to regenerate because the catalyst is enclosed within it, and because the cell regenerates, the catalyst remains captive within it.

Since the pioneering research by (Von Neumann, 1966; Gardner, 1970; Langton, 1984), researchers have gone on to study richer patterns, introducing biochemical mechanisms, physical mechanisms and genomic elements (Dittrich, Ziegler, & Banzhaf, 2001; Madina, Ono, & Ikegami, 2003; Watanabe, Koizumi, Kishi, Nakamura, Kobayashi, Kazuno, Suzuki, Asada, & Tominaga, 2007; Hutton, 2007). Both fields of research and reported results have thus become much more diverse. Consequently, in this section, we shall deal only with the research which explicitly mentions autopoiesis.

Following on from the analysis put forward by Barry McMullin in (Mc Mullin, 2004), we have organized the different approaches into three categories:

1- The study of the dynamics of basic principles in minimalist models aiming at a mathematical analysis of the system's viability (Bourgine & Stewart, 2004; Ruiz-Mirazo & Mavelli, 2008). This analysis is conducted using stochastic differential equations. These equations imitate the way in which concentrations of the elements forming the system evolve and establish stability criteria for these elements. For these approaches, the viability of the system represents its ability to keep its concentration stable when under strain from external forces. The topology of the system cannot be manipulated via these systems. For example, the position of the membrane of the tessellation automaton is predefined in (Bourgine & Stewart, 2004).

It follows that the notions of interior and exterior are themselves implicit. However, this topological distribution plays an important role in the principle of autopoesis and in evolving phenomena such as distortion, which cannot be replicated.

2- The study of the plasticity of configurations which can be preserved during disruptions or which enable the minimal action of an artificial entity (Beer, 2004; Moreno, Etxeberria, & Umerez, 2008). These studies involve the configurations of different cellular automata. This time, the topological elements can be simulated using this type of automaton. The viability of this approach depends on the preservation or evolution of a shape inscribed on the grid. Whereas (Beer, 2004) addresses the configurations of the game of life, (Moreno et al., 2008) develops (Varela et al., 1974)'s initial automaton, giving it the ability to move around under the influence of a flow of substrate on the grid. They also demonstrate the influence of the automaton's specifications on the ability of the cell to move around.

3- The study of the emergence of autopoietic behavior (Beurier, Simonin, & Ferber, 2002). The authors base their research on the notion of multiple emergences using a situated multi-agent system. Viability is summarized as the maintenance of the emergent process. Different agents positioned on a grid mutually attract or repel one another according to pre-defined rules and the virtual pheromones that they diffuse onto that grid. They can also change their "nature" (this nature being represented by a variable), depending on the state of their surroundings. This model exhibits properties of autopoietic systems: membraneionic auto-organization of the system, preferential interaction between the elements of this auto-organization, and finally the ability to withstand disruptions and to regenerate the system should it become damaged.

*The problem of co-evolution*

The possibility of co-evolution for each of these approaches is linked to the difference of opinion surrounding the notion of viability. This clearly illustrates the variety of different ways in which the autopoiesis principle can be interpreted. It also raises the issue of status in the "topological and physical nature" of autopoietic principles. For example, the notion of the frontier is intuitively topological but can become completely abstract in a digital phenomenal domain. Nevertheless, the first category of approaches does not follow the causality of the entity's internal mechanisms. These models therefore do not convey the granularity necessary to be able introduce the equivalent of a membraneionic distortion or an interaction with an environment whose characteristics would evolve. To do so would involve using a simulation, integrating the physical constraints of collision and movement. In (Manac'h & De Loor,2007), we presented the simulation of one such model based on agents situated in a continuous three-dimensional universe (see figure 2). These simulations show the extent to which it is difficult to recreate the theoretical results of stabilization demonstrated in simplified mathematical analyses. Similar work introducing physical parameters such as pressure or hydrophobia have been put forward by (Madina et al., 2003). This is a first step towards integrating the distortion and thus the evolution of the cell.

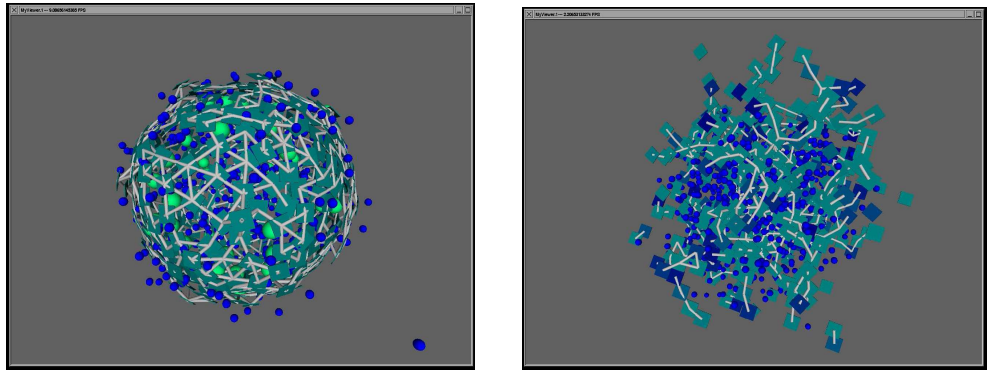

**Fig. 2** A flexible three-dimensional model of a tessellation automaton (and on the right, of its breakdown). The membrane cells (in green) are connected by springs (in gray) which disintegrate over time. However, the substrate crossing the cell (in blue), can regenerate those links when in the presence of a catalyst enclosed within the cell. After a certain amount of time, the impacts cased by the collisions deform the cell which, in the end, disintegrates (Manac'h & De Loor, 2007).

The second category explicitly introduces the evolution of the form. However, the discrete nature of cellular automata as described by (Beer, 2004) means that change of form are abrupt. The system is therefore fragile as it is sensitive to an evolving environment. Furthermore, it is the preservation of form over time that is considered proof of viability. In a context such as this, it is impossible to achieve irreversibility. It must be noted that the problem does not exist for (Moreno et al., 2008)'s approach, which could more easily tend towards co-evolution. The third category explicitly concerns emergence supported by internal rules and variables. Research is still required in order to enable these rules to evolve according to their environment.

In more general terms, to achieve co-evolution these approaches must address the possibility of acting towards and modifying the environment which, in turn, could modify the autopoietic entity. In order to do so, the roles of the environment and of the modification must be explicitly incorporated. Nevertheless, the main issue in terms of enaction based artificial intelligence, which remains in the background of this approach, is still the relevance of this detail and of the phenomenal nature of the autopoiesis principle itself. Precise biological considerations are not, by definition, necessary if the principles put forward by (Froese & Ziemke, 2009) can exist at the heart of an artificial model. Artificial dynamic cognition was developed based on considerations much like these.

**Autonomy through action: Artificial dynamic cognition**

Being linked to evolutionary robotics (Pfeifer & Scheier, 1999; Nolfi & Floreano, 2000), artificial dynamic cognition explicitly addresses the capacity of sensorimotor loops with regards to the preservation of an agent's viability (Beer, 2000; Daucé, 2002; Harvey et al., 2005). It is often claimed that it is associated with enaction even if, erring on the side of caution, the term "Enactive Artificial Intelligence" is not explicitly mentioned. For example, (Rohde & Di Paolo, 2006) suggest, that at least for now, evolutionary robotics might simply serve to study the hypotheses of cognitive science. In order to do so, they propose to concentrate on specific aspects of natural behavior so as to reduce the complexity of the problem as a whole. However, this would mean that it would be necessary to take precautions in the conception of such a reduced operation as complexity, dynamicity and plasticity must all prevail. This is one of the main challenges of this approach.

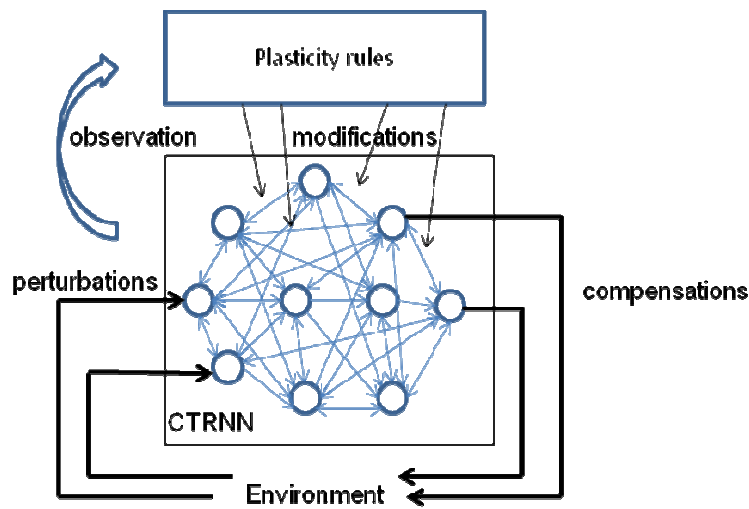

**Fig. 3** The network of neurons is recurrent and generates oscillations which are disrupted by the environment. Plasticity consists of altering the specifications of the differential equations using different criteria (ultra-stability, Hebb's laws, etc.).

The physiochemical phenomenal domain addressed by the approaches simulating autopoiesis is not discussed here, so that we might concentrate on the dynamic neuronal domain of a complete agent. The notion of viability will therefore undergo a change of perspective. The model of reference here is the Continuous Time Recurrent Neural Network (CTRNN) (Beer & Gallagher, 1992), which originates from the theory of dynamical systems (Strogatz, 1994). (Funahashi & Nakamura, 1993) highlights the advantage of being able to estimate the majority of families of dynamical systems. A network like this has chaotic dynamic behavior endowed with attractors (Bersini & Sener, 2002). In concrete terms, all of the nodes are interconnected and the output value of each one is defined by a differential equation. The parameters for these equations are defined using genetic algorithms which select and improve the right solutions according to Darwinian metaphor. In the rest of the article we shall address phylogenetic approaches. The outflows of the nodes have an oscillatory pattern. A very small proportion of arcs are linked to the agent's sensors or actuators. The difficulty is to develop networks which will enable these sensorimotor loops to auto-adapt as they gain experience. Research by (Beer & Gallagher, 1992) demonstrates an adaptation like this by giving the example of a network whose dynamic compensates for a modification of the robot's body. To do so, the genetic algorithm preselected individuals functioning with and without the modification. The network was also pre-structured and not completely recurrent. There is no approximator so universal as the unstructured CTRNN model. In order to overcome these limitations, E. Di Paolo proposes to render the network of neurons plastic so as to allow a modification of the connections' characteristics as the robot gains experience. Different plasticities may be used. Homeostatic plasticity, that which is closest to enaction, is based on the notion of ultra-stability by Ashby (Ashby, 1960). It consists of setting up a stabilization loop which will modify the network arcs involved in the over- or under-activity of neurons. In comparison with the biological conditions of the organism, maintaining these values within an interval represents a condition of the viability of the network such as maintaining a certain temperature or blood-sugar level. Hebbian plasticity consists of adjusting the weight of network arcs according to the correlation or non-correlation of the activities of the nodes which they link together. In both cases, the rules of plasticity are defined by genetic algorithms. (Wood & Di Paolo, 2007) compare these techniques, complicating homeostatic behavior by defining the zones of stable homeostatic functioning designed for precise activities (Iizuka & Di Paolo, 2007).

The general pertinence of these approaches has been demonstrated by reproducing numerous experiments, often inspired by psychology. For example, (Di Paolo, 2000) explains the architecture used to give a robot the ability to make up for a visual inversion when following a target (inversion of the robot's sensors). What is remarkable is that, when the sensors are inverted, the rules of plasticity are activated and the robot is able to behave as it should, even though these rules have never been phylogenetically learnt in such conditions. Here, phylogenesis has allowed the preservation of adequate internal dynamic behavior for the viability of the system, even if the sensorimotor loops must be modified accordingly. Another remarkable factor is that the longer the functioning period in a particular mode, the longer the re-adaptation will be, thus supporting Ashby's theory and the psychological approach. Using other experiments, (Harvey et al., 2005) demonstrated that these networks possess the ability to remember, and (Wood & Di Paolo, 2007) highlight behaviors also observed during psychological experiments with children.

*Problems for co-evolution*

Other researches involving the plastic evolution of neuronal networks are presented in evolutionary robotics (Floreano & Urzelai, 2000). However, we have presented the findings of E.A. Di Paolo's team, as they are particularly representative of enactive inspiration and insist upon the system's agentivity. Plasticity also enables the system to auto-adapt to its environment using the principle of ultra-stability, which is fundamental to this domain (Ikegami & Suzuki, 2008). However, even if the action of the robot is followed, the environment is not altered in the irreversible sense of the word mentioned in section2. The robot moves, but does not undergo an irreversible modification in its environment. For example, if the sensors of the phototaxic robot are inverted, the plasticity of the neuronal network will enable it to behave correctly. In theory, if we return the sensors to their initial position, the configuration of the neuronal network should return to its initial state and the experiment could be repeated as many times as we like, without any major changes occurring between them except, perhaps, readaptation time. In other terms, the visual inversion experiment does not irreversibly alter the phototaxic robot. Its experience will not have taught it anything, nor changed it in any way. Therefore, the saying "one never forgets", is not supported by a model such as this. Knowledge is stored in the network's dynamics, but the following stage, in which the entity retains and remembers that knowledge so that the system might use it in the future, is missing. The difficulty is in finding the "essential variables" associated with rules which could enable a more radical evolution than this. The principle of ultra-stability alone does not give access to that of irreversibility, at least in simplified models. As (Ikegami & Suzuki, 2008) suggest, the entity must also be subject to evolution. In fact, the evolutionary approaches are also faced with the problem of phylogenetic/ontogenetic articulation, which seems to be extremely difficult to resolve.

# Propositions: Toward co-evolution with humans in the loop

### Positioning

We shall now go on to present a proposition that aims to push back the limits previously identified here so as to enable an EBAI to refine its agentivity by means of more complex co-evolution. This proposal is based on the following arguments involving irreversibility, ontogenesis and sense-making.

*The problem of irreversibility*

The irreversibility of co-evolution is often overlooked as the evolution of the environment, which follows the actions of the agent, is neglected in favor of initiating an adaptivity to external changes, i.e. those which do not follow the actions of the agent itself. We suggest that the agent should actively modify an environment which, in turn, should also evolve. This principle is based on research suggesting that an entity's environment is made up of other similar entities (Nolfi & Floreano, 1998; Floreano, Mitri, Magnenat, & Keller, 2007). It is a mechanism such as this which must be set up for the preceding entities. In the following section we shall present our arguments to support the hypothesis that this is not sufficient to control this co-evolution nor to enable it to access sense-making which might be relevant to humans. First, we shall try to complete the principles suggested by (Froese & Ziemke, 2009) for the constitution of an agent from an enactive perspective, by a "principe of irreversibility".

- **EBAI irreversibility design principle:** an artificial agent must have the ability to actively regulate its structural coupling, depending on its viability constraints, with an environment which it modifies and for which certain modifications are irreversible.

This implies that it is possible that, as a result of an action, the agent's perception of its environment may be altered in such a way that it will never again perceive that environment in the same way. The fact that this only involves certain modifications and not all of them thus enables the agent to stabilize its coupling, which cannot be done in an environment which is too flexible. One difficulty is thus to find the balance between sufficient resistance for it to be able to remember the interactions, an *"en habitus deposition"* (Husserl, 1938), and sufficient plasticity for it to be able to evolve.

*The problem of ontogenesis*

Even if the modeled agents are complex in the sense that we call upon the notion of emergence in order to characterize their general behavior, their ontogenesis can be considered relatively simple. Either the principles of autopoiesis and viability are the sole focus of attention, to the detriment of the evolution of these principles or, the ontogenesis of the agent is defined using an evolutionary approach. However, the Darwinian inspiration behind the evolutionary approach is not compatible with an explanation of ontogenesis as it evaluated a whole agent. The agent is ready to function and fulfill the task that it has been selected for. That being said, if we want to progress in terms of capacity, and to broaden the cognitive domain of artificial agents, we must take into account the fact that the more complex agents are, the greater the ontogenetic component of their behavior compared to the phylogenetic component. Furthermore, as they develop, the influence of the environment becomes superior to the influence of genetic predetermination (Piaget, 1975; Vaario, 1994). From an enactive perspective, evolution is considered more as a process of auto-organization than a process of adaptation. It is therefore important to distinguish between an auto-adaptive system and a system which learns (Floreano & Urzelai, 2000). For example, in robotics, it is necessary to express evolutionary research differently so that it does not rely on the selection of agents capable of fulfilling a task or of adapting to a changing environment, but rather on a selection of agents capable of "adapting their adaptation" to that of the other and thus cope with new environments. This is debatable, as we could argue that the behavioral creativity of natural organisms is inherited from the adaptation characteristics selected throughout their phylogenesis. It remains nonetheless true that every organism's past conditions both its identity and what it will become, and especially so in the case of organisms with highly developed cognitive abilities (Piaget, 1975). Even if the aforementioned research shows that the principle of ultra-stability supports this argument, one important issue still needs to be addressed: that of the generalization of ontogenetic development principles. This problem is so tricky that we suggest associating evolutionary approaches with guided online learning, during ontogenesis. Here, we fall under a Vygotskian perspective according to which training constitutes a systematic enterprise which fundamentally restructures all of the behavioral functions; it can be defined as the artificial control of the natural development process (Vygotsky, 1986). Now is a good moment to refer back to the biological world from which, generally speaking, we can deduce that the greater an organism's cognitive capacities, the greater the need for guidance in the early stages of its life. This may require the use of a different kind of model of plasticity, for example morphological plasticity of the configuration of the system itself so that it might increase and specialize selected components as it gains experience. The problem of explaining these principles and proposing models, techniques and processes capable of recognizing them is thus raised.

*The problem of sense-making*

Let us imagine that the previous step has been achieved and that we know how to obtain an artificial system capable of co-evolution. Let us also imagine that we could imitate the environment of such a system in the same way as the system itself. There would be a co-evolution of these two entities. Both systems could engage themselves along "uncontrollable natural derivations". Enaction considers that a subject's world is simply the result of its actions on its senses. Thus, the presence of sensorimotor invariants evolving at the heart of an artificial system is the machine's equivalent of "virtual sense-making" in the virtual own-world. What would this sense-making represent for an artificial system co-evolving with another artificial system? We must be wary of anthropomorphism, which is inappropriate here as the construction of meaning and sense for such machines cannot be compared to those of humans. We argue that meaning; coherent within the perspective of Man using the machine, and evolving from the cooperation between Man and machine, can only emerge through interactions with a human observer. Otherwise we will find ourselves faced with machines resembling patterns created by fractal evolutionary algorithms. They would be extremely complex and seem well organized, but would be incapable of forming social and shared meaning. This by no means leads us to question the value of experiments in evolutionary robotics for the understanding of fundamental cognitive principles, but rather to attempt to address the problem of sense-making. We must nevertheless take precautions, keeping in mind the potential impossibility of attaining such knowledge, just as (Rohde & Stewart, 2008) argue for the notion of autonomy. We simply wish to explore the leads which might enable us to come closer to one of the aims of artificial intelligence: the confrontation of a human user and a machine (Turing,1950). We hypothesize that, from an enactive perspective, one relevant approach would be to explore the sensorimotor confrontation between Man and machine. In this context, we believe that Man must feel the "presence" of the machine which expresses itself by a sensorimotor resistance in order to construct meaning about itself. This idea of a presence, much like the Turing test, evaluates itself subjectively. This has notably been studied in the domain of virtual reality (Auvray, Hanneton, Lenay, & O Regan, 2005; Sanchez & Slater, 2005; Brogni, Vinayagamoorthy, Steed, & Slater, 2007) and enables us to link phenomenology and Enaction-Based Artificial Intelligence. We therefore make the hypothesis that a presence test could be to Enaction-Based Artificial Intelligence what the Turing test is to the computational approach to AI. An EBAI compatible with this presence test must be in sensorimotor interaction with Man in order to coordinate its actions with those of the machine, which in turn could guide and learn from it so that together they might construct "interaction meanings".

(De Jaegher & Di Paolo, 2007) comment on the participatory aspect and on coordination as a basis for the construction of meaning in an enactive perspective. The actions of the other are as important as the actions of a subject in contributing to the enaction of its knowledge. Thus, we argue that the human's participation in this co-evolution will enable both he and the machine to create meaning. If Man is not part of this loop, from his point of view there is no intelligent system. Inversely, with his participation, the coupling causes an own-world to emerge for the user. This raises the issue of the mode of interaction between Man and machine, which we shall address in section 4.2.

*Summary of our proposals*

To clarify our remarks, our proposals are summarized in the following paragraph:

- **Proposal 1:** To overcome the problem of irreversibility, we propose to add a principle obliging the agent to actively modify an environment which would also be evolving.

- **Proposal 2:** In order to overcome the issue of the complexity of ontogenesis, we propose the introduction of interactive guidance for the agent throughout its ontogenesis so as to leave it a memory of its interactions, as in the case of complex cognition in the animal kingdom.

- **Proposal 3:** To overcome the problem of the creation of relevant meaning in terms of the presence test, we suggest integrating humans into the loop so that a co-creation of meaning relevant to Man might also occur in the artificial system.

These three proposals should not be addressed head-on. To us, it would seem appropriate to address the evolution of the environment without considering Man's presence in the loop or even to set up interactive guidance without addressing the environment. However, for each of these stages, we must not lose sight of the ultimate necessity for these two elements in order to guide the theoretical or technical choices that must be made when designing them. The final objective is to design ontogenetic mechanisms for complex dynamical systems which will be guided by people. This objective is illustrated in figure 4. Artificial entities are complex systems enriched with ontogenetic mechanisms which guide their evolution via an "*en habitus deposition*" of their interactions. This guidance can be conducted via a simulated environment, but must include human interaction. We shall see that it must be done using enactive interfaces. The complexity of online guidance such as this leads us to imagine progressive exercises linking the evolutionary and ontogenetic approaches. We shall thus present the elements which seem relevant to the instigation of our research program. Section 4.2 addresses the issue of the interface between Man and machine, and section 4.3 addresses that of guidance and ontogenesis.

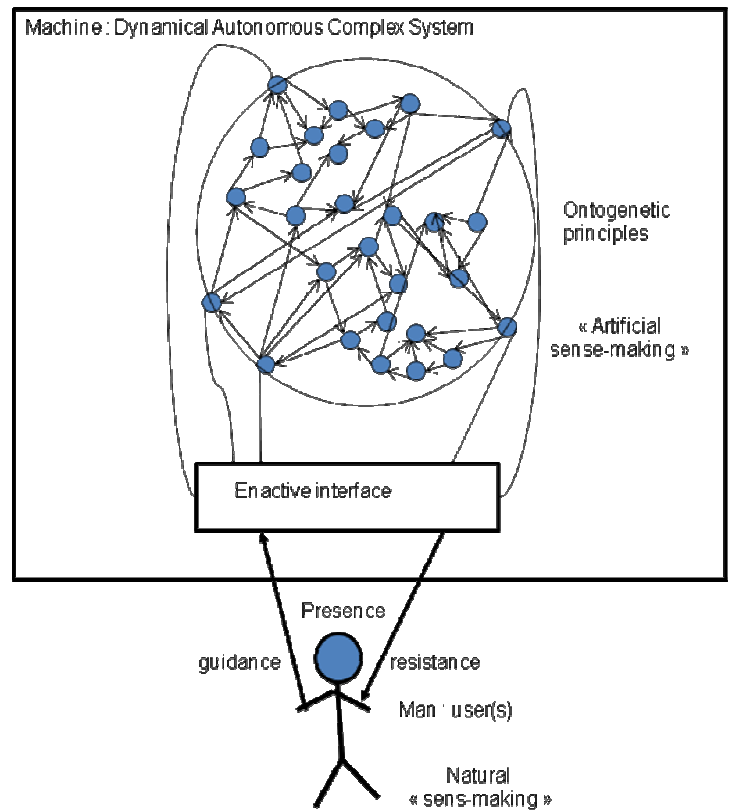

**Fig. 4** Artificial entity based on enaction metaphor.

**Interface requirements**

The interface between the system and its environment is one of the more delicate points of our proposal. Indeed, in enaction, the notion of the body as an entity able to feel and to act, originating from Merleau-Ponty, is essential and should be referred back to for an artificial system. The own-body conditions the creation of an own-world. What do own-world and body mean to an artificial entity? We must admit that, with our technique in its current state, there is a substantial difference between a machine and a living organism in terms of both body and cognition. Due to the technical implications, we are obliged to restore the separation between the cognitive element and its form which, together make up the equivalent of an own-body. The entity's"form" is actually a keyboard, a mouse, a screen, a speaker or any other device which represents the behavioral interfaces of virtual reality. It thus transforms mechanical and energetic signals into electronic signals and vice versa. The form conditions the combinations of physical and thus electronic signals. These electronic signals represent the entrances/exits of the cognitive system (for which we reiterate the temporary status). The form simply limits the possible combinations between the entrances and exits of the cognitive system. In this context, these entrances and exits are not to be considered as representations of a pre-given world, but as a means of coupling for the cognitive system and the environment. That which is technically referred to as a system's entrance or an exit point has no bearing on the notion of information but rather on dynamics. These entrances and exits are elements of the sensorimotor loops. The artificial system's ability to act thus correlates with its ability to modify the links between the entrances and exits of the cognitive system already bound within its cover. The complexity of the artificial self-world is relative to the richness of the possible advents of successive entrances/exits of the cognitive system. The more possible successive entrances and exits, the more variable, and thus more creative, the system will become. Of course, the complexity of the cover can make up for the simplicity of the cognitive system (McGeer,1990), but the opposite is also true. The nature of a sensorimotor's system, complex as it may be, is still not comparable to that of a human. The claim that the machine must have a physical body similar to ours is thus problematic (Brooks, 1991). Whatever the physical interface enabling the machine and its environment to interact, this interaction is nothing but a disturbance of the digital sensorimotor system. This is not the case for wholly embodied biological human beings who must be endowed with enactive interfaces (Luciani & Cadoz, 2007). These interfaces consist of replacing the symbolic communications (words, icons, etc.) between Man and machine with an interaction, using gestures and forces which then form "phycons". We believe that numerous types of enactive interfaces between the system and its user are possible as perception is a morphogenetic process (Gapenne, 2008). Once perception and virtual or digital action become dynamically interwoven within the machine, the technical interface can be both simple and varied. The important elements here are the presence of an uninterruptable dynamic, the absence of given symbols and the presence of evolving processes on both sides of the interface. A simple example of an interface like this is that used in minimalist experiments of the recognition and awareness of space in blind subjects (Auvray et al., 2005). For the blind subject to be able to perceive, she must be able to act and to find sensorimotor invariants. This experiment is even more interesting as (Stewart & Gapenne, 2004) has shown that these interactions can be recreated by a machine using qualitative descriptions of the experiment. Experiments such as these have led to the rethinking of the notion of virtual reality in order to bring it closer to the notion of resistance (Tisseau, 2001) and presence (Sanchez & Slater, 2005; Brogni et al., 2007; Rohde & Stewart, 2008) which we referred to earlier: whatever the chosen means of interaction, the essence of virtual reality can be identified as its ability to resist actions, to enable the user to construct meaning. Similarly, "real virtuality" could be created by an artificial system if it could negotiate its own resistance with that of its user and establish its own sensorimotor invariants. In this case, we would be confronted with an artificial sense-making comparable to that of humans.

**Guiding and explaining ontogenesis**

We have argued for the need to use models whose characteristics are irreversibly transformed through ontogenesis during the interaction, which also acts as guidance. To do so, we would need to associate learning techniques such as reinforcement (Sutton & Barto, 1998) or imitation (Mataric, 2001) with the principles of transformation and evolution. Different approaches could be used and combined.

Learning by reinforcement, which allows an entity to use its past experience to modify its behavior is used as much for symbolic connotation models (Holland & Reitman, 1978; Wilson, 1987; Butz, Goldberg, & Stolzmann, 2000; Gerard, Stolzmann, & Sigaud, 2002) as for neurocomputational approaches (Daucé, Quoy,

Cessac, Doyon, & Samuelides, 1998; Henry, Daucé, & Soula, 2007). We discuss "symbolic connotation" approaches first as they are based on discrete variables and a selection of atomic actions. However, they are not confined to using a given representation of an environment but rather to create a model of possible coupling with this environment. They are thus viable for consideration in our context. Recently, (Chandrasekharan & Stewart, 2007) have shown that it is possible to associate a network of neurons which loop back to themselves with a Q-learning type of algorithm. The functioning of this network serves as a proto-representation. The idea is that these protorepresentations act as internal epistemic structures which reflect the sensorimotor invariants learnt by experience. However, the learning process requires hundreds of simulated steps for a simple example (i.e. virtual ants foraging). This is a serious drawback for an online application of these approaches. In terms of neurocomputation, (Henry et al., 2007) proposes reinforcement learning for a network of recurrent neurons. The network's Hebbian plasticity is only activated in the presence of reward or punishment stimuli. This approach is also well adapted to our context. However, there is a difference between this approach and artificial dynamic cognition, as the experiments are not based on sensorimotor learning and the networks of neurons used are not CTRNN.

To fully address the notion of transformation, the introduction of morphogenetic principles gives the advantage of being able to access irreversibility. With the work on modeling the growth of a mutlicellular organism, we therefore return to the biological origins of cognition (Federici & Downing, 2006; Stockholm, Benchaouir, Picot, Rameau, Neildeiz, & Paldi, 2007; Neildeiz, Parisot, Vignal, Rameau, Stockholm, Picot, Allo, Le Bec, Laplace, & Paldi, 2008). Certain authors even introduce the role of the environment into this transformation (Eggenberger, 2004; Beurier et al., 2006). In robotics, it is the evolution of the body of the machine that is of interest (Dellaert & Beer, 1994; Hara & Pfeifer, 2003). Eventually, these approaches might access co-evolution in the fullest sense of the term. In the case of an EBAI, the connection which should be made is to integrate the principles of autopoiesis with those of morphogenesis so as to preserve the biological essence of an identity built up within the constraints of viability (Miller, 2003). The research pertaining to neurocomputing can be found in (Gruau, 1994; Nolfi & Parisi, 1995; G.vHornby & J.Pollack, 2002). Finally, in order to study these principles, we must rely on the formal tools adapted to the models which present the properties of multiple behavioral drifts. However, these tools are uncommon and in (Aubin, 1991)'s theory of viability, which aims to define all of the parameters of models capable maintaining their own behavior in a given area, we can observe an interesting perspective for associating the simulation's bottom-up approach with the analysis of global properties. We believe this theory to be under-used, whilst it suggests a turnaround in terms of the most common point of view in studying complex systems.

It remains that, in terms of an interaction between Man and machine, an association of the principles of reinforcement and transformation must be developed.

**Conclusion**

The aim of this paper was to analyze and define new approaches for addressing the difficulties in constructing independent artificial systems which rely on enactive metaphor. First, we brought together the notions of Enactive Artificial Intelligence and Enaction-Based Artificial Intelligence. We particularly wanted to avoid addressing certain phenomenological aspects such as the notion of first-hand experience in order to avoid any confusion with the human perspective of the paradigm. We then went on to demonstrate that the three current main approaches were confronted with the following three problems:

1. The absence of the implementation of a real co-evolution characterized by its irreversibility. To overcome this problem, we suggest that the agent should more actively modify its environment and that in turn that environment should evolve and present a certain degree of irreversibility.

2. The difficulty establishing a complex ontogenetic process "which determines its own outcome". This necessitates the modification of the phylogenesis/ontogenesis ratio that follows it so that auto-organization might prevail over auto-adaptation. As an answer to this problem, we suggest the use of interactive guidance throughout its ontogenesis, as is the case during the complication of cognition in the animal kingdom.

3. The immeasurable difference between the creation of meaning for machines and for humans. Due to this difference, the use of machines capable of exchange or social partnership with humans is rendered extremely hypothetical. To answer this problem, we propose to integrate humans into the loop so that the creation of a meaning relevant to humans might also develop within the artificial system. We also suggest that a presence test, the enactive equivalent of the Turing test from a computational angle, should be taken by the machine.

There follows the proposal to assimilate interaction between Man and machine during the ontogenetic process of an artificial entity via an enactive interface. One difficulty is thus to set up irreversible evolving mechanisms which are carried out in real time at the heart of the system. This is why we have listed the approaches that would enable us to clarify the ontogenetic transformation and to adapt them. Our perspectives tend towards the assimilation of these approaches via minimalist experiments associating evolutionary robotics with interactive guidance (Manac'h & De Loor 2009). Despite the complexity of the task to be accomplished, it seems to us that the inclusion of humans in the loop, as well as being essential *in-fine*, might help us to establish the strategies of evolution and guidance and to push back the limits of the obstacles highlighted by (Froese & Ziemke, 2009). Considering the phenomenology of interactions between Man and machine in the constitution of sensorimotor skills for humans could in fact prove an important basis for establishing analogical principles for machines. Attempts made to model and simulate such interactions by (Stewart & Gapenne, 2004) seem to us to be an important starting point. They might help us to imagine minimal experiments combining phylogenesis and ontogenesis in establishing mechanisms of "learning how to learn" via principles inspired from morphogenesis. These minimal interactions pass through simple but enactive interfaces, i.e. based on a sensorimotor dynamic. The important thing is to establish sensorimotor coupling between Man and machine and to keep in mind the praxeological, rather than the ontological, aspect of the system. The intricacies could be provided later.

These reflexions lead us to sketch some interesting perspectives in the context of interaction and virtual reality: Virtual environment constitute a good base to develop guided models capable of co-evolution. However, we must remain prudent because of the incommensurable distance between the continuous nature of the physical world which lead to the biological metabolism and the discrete nature of numerical systems. Numerical and natural worlds are based on two different phenomenal domains and the later is tremendously more complex that the former. Nevertheless, it doesn't prevent the possibility to bring forth a world *into* a dynamical simulation, even if this world will be incommensurable with such of the human. The only interest for this artificial world would be in the fact that it would be constituted by the way of human-machine interaction and consequently that human might find a sense in these interactions. If it is the case, a man-machine common sense might be co-constituted. To do that, we must imagine experimentations easy enough to be supported by actual artificial models but also representative for a human in term of co-constitutive interaction. Artistic creation seems to be favorable to following this way.

This by no means aims to disqualify the interest of approaches which do not include humans in the loop, which progress more quickly in terms of understanding internal mechanisms using artificial life and evolutionary or coevolutionary robotics. However, we here limit our research, having presented what seems to us to be the most important approaches. These thoughts are a result of our work on the necessary coupling between Man and machine for the co-construction of knowledge (Parenthon & Tisseau, 2005; Desmeulles,Querrec, Redou, Kerdlo, Misery, Rodin, & Tisseau, 2006; Favier & De Loor, 2006; De Loor, Bénard, & Bossard, 2008). This is thus a challenge for software engineering which must consider the "experience of the machine" and of its interactions, as well as those of the user. It is also a challenge for theoretical artificial intelligence which must integrate interaction at the heart of its models as suggested by (Goldin & Wegner, 2008).